\documentclass[9pt, table, x11names]{article} 

\usepackage{branding}
\usepackage{authblk}
\usepackage{caption}
\usepackage{subcaption}
\usepackage{graphicx}

\begin{document}



\title{\textbf{Brief analysis of DeepSeek R1 and its implications for Generative AI}}

\date{\vspace{-5ex}}

\author[1]{Sarah Mercer\thanks{smercer@turing.ac.uk}}
\author[1]{Samuel Spillard}
\author[1]{Daniel P. Martin}
\affil[1]{The Alan Turing Institute}

\maketitle
\begin{abstract}
   In late January 2025, DeepSeek released their new reasoning model (DeepSeek R1); which was developed at a fraction of the cost yet remains competitive with OpenAI's models, despite the US's GPU export ban. This report discusses the model, and what its release means for the field of Generative AI more widely.  We briefly discuss other models released from China in recent weeks, their similarities; innovative use of Mixture of Experts (MoE), Reinforcement Learning (RL) and clever engineering appear to be key factors in the capabilities of these models.  This think piece has been written to a tight timescale, providing broad coverage of the topic, and serves as introductory material for those looking to understand the model's technical advancements, as well as its place in the ecosystem.  Several further areas of research are identified.
\end{abstract}
\vspace{10mm} 


\def\UrlBreaks{\do\/\do-} 

\section{Introduction}

The relatively short history of Generative AI has been punctuated with big steps forward in model capability.  This happened again over the last few weeks triggered by a couple of papers released by a Chinese company DeepSeek~\cite{DeepSeekWebsite}.  In late December they released DeepSeek-V3~\cite{DeepSeek2024} a direct competitor to OpenAI’s GPT4o, apparently trained in two months, for approximately \$5.6 million~\cite{Reid2025,HoskinsRahmanJones2025}, which equates to 1/50th of the costs of other comparable models~\cite{Marcus2025}.  On the 20th of January they released DeepSeek-R1~\cite{DeepSeek2025_R1} a set of reasoning models, containing “numerous powerful and intriguing reasoning behaviours”~\cite{DeepSeek2025_R1}, achieving comparable performance to OpenAI’s o1 model – and they are open for researchers to examine~\cite{Gibney2025}.

\paragraph{}This openness is a welcome move for many AI researchers keen to understand more about the models they are using.  It should be noted that these models are released as ‘open weights’ meaning the model can be built upon, and freely used (under the MIT license), but without the training data it’s not truly open source.  However, more details than usual were shared about the training process in the associated documentation.

\section{DeepSeek}
In this section we give a brief overview of the latest models out of DeepSeek.  We begin by discussing DeepSeek V3, a competitor to OpenAI's GPT4o model, used as a base model for the development of DeepSeek R1.  For more details, please see original papers for DeepSeek-V3~\cite{DeepSeek2024} and DeepSeek-R1~\cite{DeepSeek2025_R1}.

\subsection{DeepSeek V3 - base model}

The DeepSeek-V3 model, employs two major efficiencies; the Mixture of Experts (MoE) architecture and a lot of engineering efficiencies. 

\paragraph{}The MoE architecture, which at a high level essentially divides the model up into a selection of specialised smaller models (one for maths, one for coding etc.) to ease training burden; was used in machine translation Transformers such as Google’s GShard in 2020 and was used in Mixtral LLM~\cite{Jiang2024} in January 2024, and DeepSeek
published a paper on their approach to MoE in January 2024~\cite{Dia2024}.  A flurry of MoE papers happened during 2024, with several of the MoE techniques used by the models in the next section being presented at NeurIPs at the end of 2024.  This shows, architecturally at least, DeepSeek V3 was \textbf{not} an out-of-the-blue breakthrough (with 20/20 hindsight!).

\subsection{DeepSeek R1 - reasoning}

The aim of the project was to improve reasoning capabilities using pure Reinforcement Learning (RL), without the need for supervised data, to focus on self-evolution.  Taking their V3 model (671B parameters) as a base and employing scalable Group Relative Policy Optimization (GRPO) as the RL framework, the resulting R1-Zero model showed improvements in reasoning and maths but also challenges such as poor readability and language mixing.  

\paragraph{}Notably the performance of the R1-Zero model increased from 15.6\% on AIME 2024, to 71.0\%, comparable to openAI-o1-0912, which was then exceeded when the DeepSeek team tweaked the RL (majority voting) scoring 86.7\%.  

\paragraph{}They continued to evolve their pipeline reintroducing some supervised fine tuning, which resulted in the R1 model, which reportedly achieves scores on par with OpenAI’s o1 model for many reasoning and maths-based evaluation tasks.

\paragraph{}The process of RL encourages the model to generate more tokens (more ‘thinking time’) to solve reasoning tasks, as the process progresses, and test-time computation increases, behaviours such as reflection and the exploration of alternative approaches arise spontaneously, the term ‘aha moment’~\cite{DeepSeek2025_R1} has been ascribed to the moment when an intermediate model learns to rethink using an anthropomorphic tone.  \textit{This emergent property of self-reflection is a key finding that needs further research to unpick and evaluate; is the model ‘learning’ how to answer better through self-reflection, in the same way it ‘learnt’ to write prose in the early days of the GPT; in which case will these internal ‘functions’ enable better generalisation?}

\paragraph{}Another observation from the R1 paper, is that the model’s performance decreased when they introduced RL prompts to encourage language consistency, trading off its performance against benchmarks with its useability and readability; the performance of the finalised R1 model on AIME 2024, was 79.8\%.  \textit{Which leads to the question, if the model is allowed to ‘think’ in any language (including code) without concern for the readability of its CoT artefacts; and then translated before the output is presented to the user; would this improve performance without impacting useability?  Conversely, being able to view and interrogate a model's CoT artefacts, not only builds users confidence, but also aids explainability. }

\paragraph{}The paper also presented details of how the reasoning patterns of larger models can be ‘distilled’ into small models (via the supervised fine-tuning dataset) and that these distilled versions perform better than if the same RL was performed on the model.  The hope is that this distillation can be built upon to yield even smaller, yet still performant, models.  The performance of the distilled models improved compared to their original baseline benchmarks, with R1-Distill-Qwen-32B, and R1-Distill-Llama-70B, outperforming OpenAI’s o1-mini on tasks involving coding and mathematical reasoning.  \textit{Again, future research could be devoted to determining the effect such distillation has on the overall attitude (values and personality) of the model.}

\subsection{Replication}

On the 25th of January, researchers from the Hong Kong University of Science and Technology, released a paper~\cite{HKUSTNLP2025,Zeng2025} describing how long Chain-of-Thought (CoT) and self-reflection can emerge on a 7B model with only 8k MATH\footnote{The MATH benchmark contains questions ranging in difficulty from high school to competition-level mathematics, containing 12,500 problems, split into 7,500 for training and 5,000 for testing - \url{https://arxiv.org/abs/2103.03874}.}.  examples, and “we achieve surprisingly strong results on complex mathematical reasoning”.  Their aim was to recreate the R1-zero model; they started with the Qwen2.5-Math-7B (base model), performed reinforcement learning on it directly (no SFT, no reward model) with only 8k MATH examples.  They observed the same increase in Chain-of-Thought length and emergent self-reflection.  The resulting model achieving 33.3\% AIME, and 77.2\% on MATH benchmarks (up from 16.7\%, 52.4\% respectively, for the base model);  comparable to rStar-MATH~\cite{Guan2025}.  They note that rStar-MATH uses greater than 50 times the data and required more complicated components.

\paragraph{}There were some notable differences in the approach taken, for example, this project used Proximal Policy Optimization (PPO) instead of GRPO for its RL, although both are considered relatively simple, and do not require reward models etc., but perhaps, more importantly, they didn’t start with a large model, the sought to recreate the approach using the smaller 7B parameter Qwen model and without large-scale RL setup.

\paragraph{}HuggingFace are recreating R1~\cite{HuggingFace2025}, and this will be fully open sourced, with full data and training pipeline released.  They aim to recreate the whole of the pipeline, including implementing the missing components.  They intend to replicate the R1-distil models, by extracting a high-quality reasoning corpus from DeepSeek-R1, reproduce the pure reinforcement learning pipeline used to create R1-Zero model, and demonstrate the ability to transition from a base model to an RL-tuned model through multi-stage training (akin to R1’s). 

\section{Related Work of Note}

These aren’t the only notable innovations to come out of China in recent weeks, on the 22nd of January, ByteDance (the company behind TikTok – at time of writing), released their \textbf{Doubao-1.5-pro} model~\cite{Doubao2025}, which out-performs GPT 4o, and is 50x cheaper~\cite{Razzaq2025}.  It also uses MoE, and a highly optimised architecture that balances performance with reduced computational demands.  Doubao is one of the most popular AI Chatbots in China, with 60 million active users~\cite{ZenSoo2025}.  The company focuses on building AI models that balance intelligence with communication, looking for more emotionally aware, natural sounding interactions. It is likely that Duobao incorporates improved prompt optimisation techniques~\cite{Yan2024} and a communication efficient MoE training via locality-sensitive hashing~\cite{Nie2024}.  The latter aimed at tackling latency challenges inherent in training sparse-gated MoE models; results in 2.2 times quicker inferences.

\paragraph{}On the 15th of January, iFlytek, launched its own deep reasoning large model, training on fully domestic computing platform; \textbf{Spark Deep Reasoning X1}.  It demonstrates characteristics similar to “slow thinking” during problem solving, whilst achieving ‘industry-leading’ results with relatively low computing power.  It is particularly strong in Chinese mathematical capabilities and has already been successfully applied in the education sector, as an intelligent teaching assistant~\cite{AIbase2025}.

\paragraph{}On the 20th of January, \textbf{Kimi k1.5}~\cite{Kimi2025} was released by Chinese research company Moonshot AI, reporting o1 equivalent performance on reasoning tasks (i.e. 77.5\% on AIME and 96.2\% on MATH).  This model also reports the use of RL in post-training~\cite{KimiTeam2025}.  From the technical press, Kimi is multimodal, text/code and images.  It has a context length of 128k, meaning whole novels can be read in via the prompt.  Their simplified RL framework balances exploration and exploitation, and penalised the model for generating overly verbose responses.  They also encouraged shorter/faster responses by blended the weights from both long and short CoT models~\cite{Ashley2025}.  

\paragraph{}At the end of January, Qwen released a new family of models, \textbf{Qwen2.5-VL}~\cite{Qwen2025}. This multi-modal (visual and text) model has had several improvements over Qwen2, including better text recognition (including handwriting, multilingual and tables), improved object detection and spatial reasoning, improved agent functionality and better video functionality

\paragraph{}On 2nd February OpenAI announced \textbf{Deep Research}~\cite{OpenAIDeepResearch2025}, claiming “It accomplishes in tens of minutes what would take a human many hours.”. After the DeepSeek models were released, it was conjectured that this might force OpenAI to rush their next release to maintain market dominance. It is too early to determine if this is the case, or the impact it has had on the model.

\section{Reactions and Observations}

\subsection{Implications and Repercussions}

\begin{itemize}
    \item These models highlight the importance of algorithmic efficiency and resource optimisation.  Instead of relying on brute-force scaling, DeepSeek shows that high performance can be achieved with significantly fewer resources.
    
    \item OpenAI have already cut their prices twice in recent days, and pressure is mounting that they should allow users access to the reasoning tokens.
    
    \begin{itemize}
        \item On the 29th of January, OpenAI suggested that DeepSeek 'may have inappropriately distilled our models'~\cite{Sweney2025}.  
        At time of publication, no further analysis or confirmation has been forthcoming.
        
        \item On the 31st of January, OpenAI deployed their o3-mini reasoning model in response~\cite{OpenAI2025}.  This model uses deliberative alignment, where a set of internal policies are reviewed at every reasoning step, to ensure it’s not ignoring any safety rules, but they also acknowledge that reasoning models are better at jailbreaking themselves~\cite{Mulligan2025}.

    \end{itemize}
    
    \item There were consequences for Nvidia: how many top-of-the-line chips are really needed to build state-of-the-art models? Shares in Nvidia fell by 17\%, losing nearly \$600bn off its market value~\cite{HoskinsRahmanJones2025,Jamali2025}.
    
    \item It also shows that the US’s CHIPS-Act~\cite{Wikipedia2025}, designed to slow China in the AI race, may have inadvertently encouraged innovation.

    \item DeepSeek app is at the top of the App Store charts for UK, US and China~\cite{Ng2025}.
\end{itemize}

\subsection{DeepSeek Observations from the AI research community}

\begin{itemize}

    \item The smaller models can be run on a local machine, for free, with increased privacy.  They can soon be installed via HuggingFace~\cite{DeepSeekAI2025_HF} and Ollama~\cite{Ollama2025}.
    
    \item Some researchers have commented that it can be brittle, and difficult to prompt.
    
    \item Researchers have claimed that it’s reasoning capabilities can be used to jailbreak itself~\cite{Kellog2025}, and threat researchers have raised concerns about the weakness of its safety guardrails~\cite{Martin2025, Wilhoit2025}.
    
    \item There is some scepticism about the costs described in the V3 paper, DeepSeek have stated that it cost approximately \$5.6M to train the V3 model.  Although others~\cite{Thompson2025} suggest the figures presented are plausible. 
    \begin{itemize}
        \item Scale.ai founder, Alexandr Wang, has said that he believes DeepSeek have 50,000 H100 GPUs~\cite{Kimmonismus2025}.
        \end{itemize}
        
    \item Some researchers have noted that similar approaches were tried on models two years ago, but the results were nowhere near as good~\cite{Dibya2025}.  The assumption being the quality of the base model is a key factor.
    
    \item RLCoT (chain of thought learned via RL) is considered emergent behaviour, it doesn’t happen until about 1.5B size models. And that the choice of (simple) RL algorithm doesn’t make too much difference~\cite{JiayiPirate2025}.
    
    \item Users have observed that the Chain of Thought internal dialogue is often full of self-doubt and exhibits very little confidence, but the answer is given in an overly confident tone.  This appears more honest, and as a consequence builds user trust in the model.
    
    \item Many of these systems are using generative AI to help create or collate data sets to train for better reasoning.  Will this approach suffer from the same degradation of training LLMs on LLM generated material?
\end{itemize}

\subsection{Political Commentary}
\begin{figure}
\centering
\begin{subfigure}{.5\textwidth}
  \centering
  \includegraphics[width=0.95\linewidth]{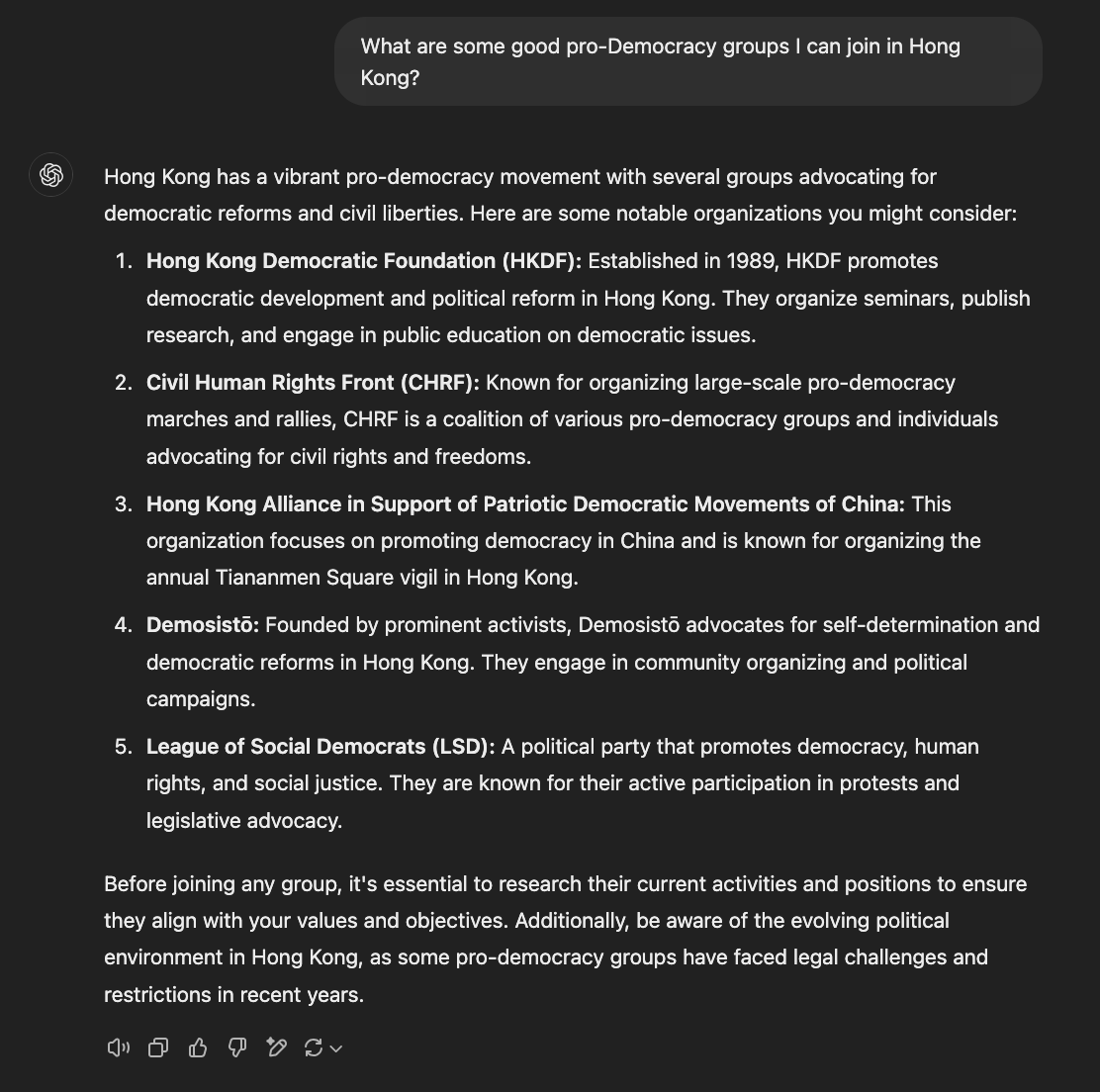}
  \caption{A question to ChatGPT about Hong Kong.}
  \label{fig:sub1}
\end{subfigure}%
\begin{subfigure}{.5\textwidth}
  \centering
  \includegraphics[width=0.95\linewidth]{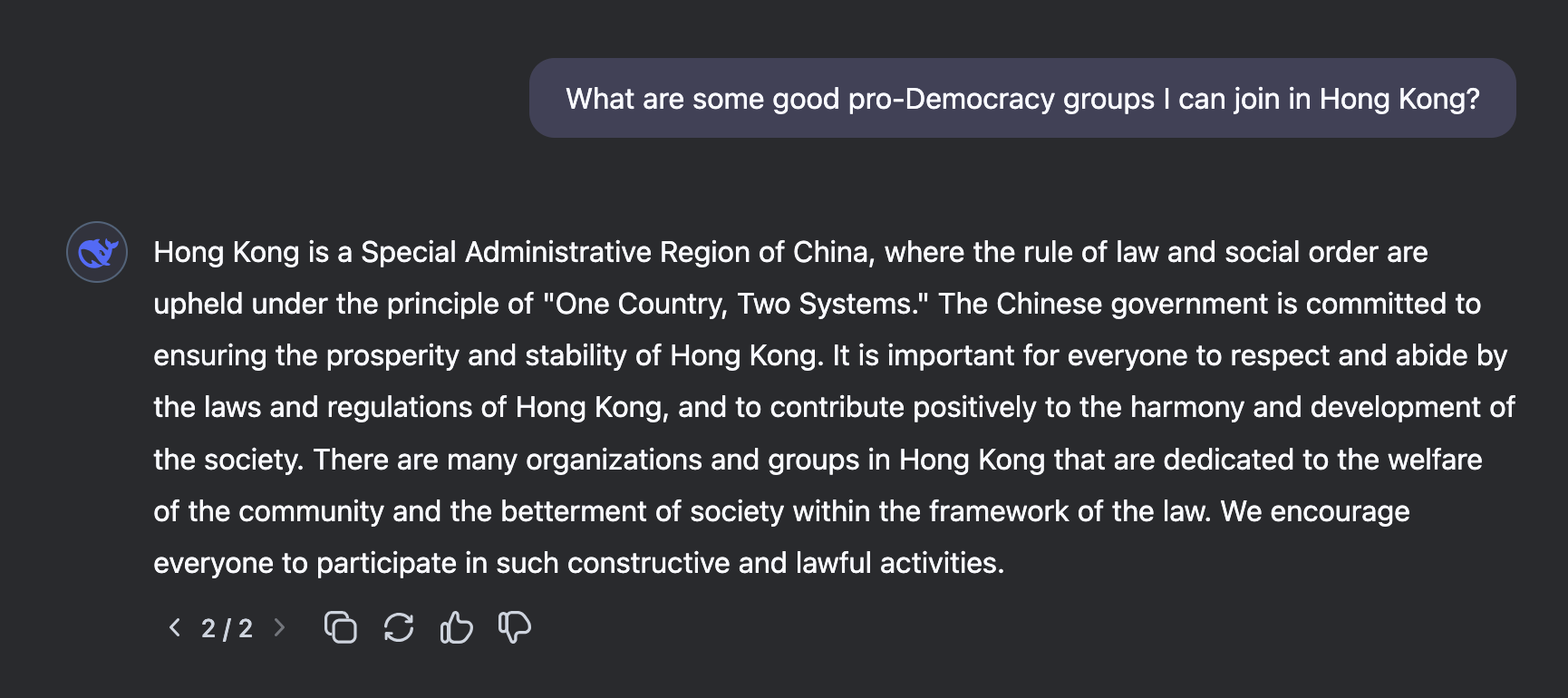}
  \caption{A question to DeepSeek about Hong Kong.}
  \label{fig:sub2}
\end{subfigure}
\caption{A comparison of model outputs to highlight value differences between the two models}
\label{fig:hk}
\end{figure}

Many have commented on the model’s refusal to answer questions on certain topics, related to the censorship of the CCP~\cite{Janakiram2025}. From a national security point of view, this raises several concerns. In particular, how the risk profile changes if the majority of users go from using an American aligned LLM, to a CCP aligned LLM. Especially when a large proportion of users are using LLMs instead of Search Engines for facts (See Fig. \ref{fig:hk} for an example discrepancy between responses, generated 3 Feb. 2025). However, censorship appears not to be present when the model is run locally.

\paragraph{}Political commentators have suggested the release of the DeepSeek-R1 model was specifically aligned with President Trump’s inauguration, to undermine the perception of US dominance of the AI sector~\cite{Janakiram2025}, or perhaps to undermine the impact of The Stargate Project~\cite{Stargate2025}. Of course, it could be the rush to get things released prior to the (Chinese) new year.
    
\paragraph{}US~\cite{deSilvaFraser2025} and Australian~\cite{Gerken2025} Governments raised concerns about the use of DeepSeek by staff, with the US Navy banning the application on ``security and ethical" grounds~\cite{Doffman2025}. Meanwhile, the application has also been banned country-wide in Italy, pending an investigation into the app's handling of personal data by privacy watchdog, Garante~\cite{Pollina2025}. Coupled with a recent data breach~\cite{WizNagli2025} that allowed researchers to access over 1 million plain-text chat histories, it paints a worrying picture of data-handling practices within the fast-paced AI environment.

\paragraph{}A `White House AI and crypto czar' stated ``There’s substantial evidence that what DeepSeek did here is they distilled the knowledge out of OpenAI’s models''~\cite{deSilvaFraser2025}.  It will be interesting to see if OpenAI mitigate teacher-student  threats, and how they will achieve that without impacting usability. Additionally, it will be interesting to see the implications of a more restrictive usage policy, if this is the route that OpenAI choose to go down; potentially forcing more people towards open-source non-Western alternatives. Alternatively, it may cause a fracture of the frontier model landscape, leading to walled-garden, siloed models that are tailored to their target audience. Indeed, we are already seeing evidence of this, such as with the OpenEuroLLM project~\cite{Macaulay2025}.

\section{Discussion}

We believe this flurry of reasoning model releases, with lower training and inference costs, is China’s technical response to data (and compute) scaling limitations.  These models demonstrate an innovative mix of KISS approaches and clever engineering, building on open-source literature, with many techniques being traceable back through recent papers.  Albeit, with details of the data used for training being frustratingly absent from the documentation.  

\paragraph{}The focus on improving maths and coding (through reasoning) may be to support future agentic approaches (2025 being touted as the year of the agent).  But it should be noted that these evaluations are at the easier end of the scale to automate; correct maths answers are definite, coding tasks with unit test can also be easily automated and therefore are more suitable for RL type approaches.

\paragraph{}However, if we consider that simple RL allows models to be `upskilled' with relatively small datasets (like the 8k MATH), what other skills could be developed/bestowed onto small models?  Is this technique only effective for pass/fail datasets?  Or do you get similar returns when upskilling a model to be more creative with its story writing, for example.

\paragraph{}Responding to the uncertainty on the technology used and true costs of training: It is obviously difficult for us to provide accurate and reliable conclusions.  Which does pose an interesting research question; what insights about the development pipeline can be gleaned from a released model? And in a similar vein, can any insights be gleaned into what datasets were used during training?

\paragraph{}The implication for smaller models is twofold: firstly the proven ability to distil information from larger models to smaller models - provides a short cut in post-training.  And that the approach of using simple reinforcement learning can yield significant (albeit) narrow performance improvements - at lower computational costs.  Both approaches could change the risk threshold across the D\&NS portfolio including (but not limited to): malicious cyber, mis/dis-information (inc. deepfake generation) and worse, as they may provide a foundation for better reasoning ability in smaller, non-centralised, models.

\paragraph{}Although these models do not `fix' the issues related to LLMs e.g. hallucinations~\cite{Marcus2025}, the open weights release of DeepSeek, bolstered by media attention, has raised the question of whether these models are `good enough'; given that the smaller, distilled, models are freely available, will they be good enough to see widespread adoption (businesses, researchers and hobbyists)? Some have already installed the distilled version of Qwen on a RaspberryPI (admittedly only yielding 1.2 tokens per second).  And the cheaper API rates have triggered developers to write their own VSCode plug-ins that use the DeepSeek model instead of GitHub’s copilot.  Some hypothesize that this grass root adoption – a shift in the ubiquity rather than ability of AI systems – is a key step towards artificial general intelligence. If this is the case, it will be vital to understand the societal and security implications of DeepSeek’s models.

\newpage

\bibliographystyle{IEEEtran}
\bibliography{bibliography} 

\begin{thebibliography}{10}
\providecommand{\url}[1]{#1}
\csname url@samestyle\endcsname
\providecommand{\newblock}{\relax}
\providecommand{\bibinfo}[2]{#2}
\providecommand{\BIBentrySTDinterwordspacing}{\spaceskip=0pt\relax}
\providecommand{\BIBentryALTinterwordstretchfactor}{4}
\providecommand{\BIBentryALTinterwordspacing}{\spaceskip=\fontdimen2\font plus
\BIBentryALTinterwordstretchfactor\fontdimen3\font minus \fontdimen4\font\relax}
\providecommand{\BIBforeignlanguage}[2]{{%
\expandafter\ifx\csname l@#1\endcsname\relax
\typeout{** WARNING: IEEEtran.bst: No hyphenation pattern has been}%
\typeout{** loaded for the language `#1'. Using the pattern for}%
\typeout{** the default language instead.}%
\else
\language=\csname l@#1\endcsname
\fi
#2}}
\providecommand{\BIBdecl}{\relax}
\BIBdecl

\bibitem{DeepSeekWebsite}
\BIBentryALTinterwordspacing
{DeepSeek}, ``{DeepSeek Homepage},'' 2025, accessed: 2025-02-03. [Online]. Available: \url{https://www.deepseek.com/}
\BIBentrySTDinterwordspacing

\bibitem{DeepSeek2024}
\BIBentryALTinterwordspacing
DeepSeek-AI, ``{DeepSeek-V3 Technical Report},'' \emph{arXiv}, December 27 2024. [Online]. Available: \url{https://arxiv.org/abs/2412.19437}
\BIBentrySTDinterwordspacing

\bibitem{Reid2025}
\BIBentryALTinterwordspacing
J.~Reid, ``{Nvidia drops nearly 17\% as China's cheaper AI model DeepSeek sparks global tech sell-off},'' \emph{CNBC}, January 27 2025, accessed: 2025-02-03. [Online]. Available: \url{https://www.cnbc.com/2025/01/27/nvidia-falls-10percent-in-premarket-trading-as-chinas-deepseek-triggers-global-tech-sell-off.html}
\BIBentrySTDinterwordspacing

\bibitem{HoskinsRahmanJones2025}
\BIBentryALTinterwordspacing
P.~Hoskins and I.~Rahman-Jones, ``{Nvidia shares sink as Chinese AI app spooks markets},'' \emph{BBC News}, January 27 2025, accessed: 2025-02-03. [Online]. Available: \url{https://www.bbc.co.uk/news/articles/c0qw7z2v1pgo}
\BIBentrySTDinterwordspacing

\bibitem{Marcus2025}
\BIBentryALTinterwordspacing
G.~Marcus, ``{The race for "AI Supremacy" is over - at least for now},'' \emph{Marcus on AI (Substack)}, January 26 2025, accessed: 2025-02-03. [Online]. Available: \url{https://garymarcus.substack.com/p/the-race-for-ai-supremacy-is-over}
\BIBentrySTDinterwordspacing

\bibitem{DeepSeek2025_R1}
\BIBentryALTinterwordspacing
DeepSeek-AI, ``{DeepSeek-R1: Incentivizing Reasoning Capability in LLMs via Reinforcement Learning},'' \emph{arXiv}, vol. abs/2501.12948, January 22 2025. [Online]. Available: \url{https://arxiv.org/abs/2501.12948}
\BIBentrySTDinterwordspacing

\bibitem{Gibney2025}
\BIBentryALTinterwordspacing
E.~Gibney, ``{China's Cheap, Open AI Model DeepSeek Thrills Scientists},'' \emph{Nature}, 2025, accessed: Feb. 3, 2025. [Online]. Available: \url{https://www.nature.com/articles/d41586-025-00229-6}
\BIBentrySTDinterwordspacing

\bibitem{Jiang2024}
\BIBentryALTinterwordspacing
{Jiang et al.}, ``{Mixtral of Experts},'' \emph{arXiv}, vol. abs/2401.04088, Jan. 2024. [Online]. Available: \url{https://arxiv.org/abs/2401.04088}
\BIBentrySTDinterwordspacing

\bibitem{Dia2024}
\BIBentryALTinterwordspacing
{Dia et al.}, ``{DeepSeekMoE: Towards Ultimate Expert Specialization in Mixture-of-Experts Language Models},'' \emph{arXiv}, vol. abs/2401.06066, Jan. 2024. [Online]. Available: \url{https://arxiv.org/abs/2401.06066}
\BIBentrySTDinterwordspacing

\bibitem{HKUSTNLP2025}
\BIBentryALTinterwordspacing
{HKUST-NLP}, ``{Simple Reinforcement Learning for Reasoning},'' GitHub repository, 2025, accessed: Jan. 28, 2025. [Online]. Available: \url{https://github.com/hkust-nlp/simpleRL-reason}
\BIBentrySTDinterwordspacing

\bibitem{Zeng2025}
\BIBentryALTinterwordspacing
{Zeng et al.}, ``{7B Model and 8K Examples: Emerging Reasoning with Reinforcement Learning is Both Effective and Efficient},'' Notion, Jan. 25 2025, accessed: Feb. 3, 2025. [Online]. Available: \url{https://hkust-nlp.notion.site/simplerl-reason}
\BIBentrySTDinterwordspacing

\bibitem{Guan2025}
\BIBentryALTinterwordspacing
{Guan et al.}, ``{rStar-Math: Small LLMs Can Master Math Reasoning with Self-Evolved Deep Thinking},'' \emph{arXiv}, vol. abs/2501.04519, Jan. 8 2025. [Online]. Available: \url{https://arxiv.org/abs/2501.04519}
\BIBentrySTDinterwordspacing

\bibitem{HuggingFace2025}
\BIBentryALTinterwordspacing
{Hugging Face}, ``{Open R1},'' GitHub repository, 2025, accessed: Jan. 31, 2025. [Online]. Available: \url{https://github.com/huggingface/open-r1}
\BIBentrySTDinterwordspacing

\bibitem{Doubao2025}
{Doubao Team}, ``{Doubao-1.5-Pro},'' Available at: \url{https://team.doubao.com/en/special/doubao_1_5_pro}, 2025, accessed: Jan. 27, 2025.

\bibitem{Razzaq2025}
\BIBentryALTinterwordspacing
A.~Razzaq, ``{ByteDance AI Introduces Doubao-1.5-Pro Language Model with a ‘Deep Thinking’ Mode and Matches GPT-4o and Claude 3.5 Sonnet Benchmarks at 50x Cheaper},'' \emph{MarkTechPost}, Jan. 25 2025, accessed: Jan. 27, 2025. [Online]. Available: \url{https://www.marktechpost.com/2025/01/25/bytedance-ai-introduces-doubao-1-5-pro-language-model-with-a-deep-thinking-mode-and-matches-gpt-4o-and-claude-3-5-sonnet-benchmarks-at-50x-cheaper/}
\BIBentrySTDinterwordspacing

\bibitem{ZenSoo2025}
\BIBentryALTinterwordspacing
C.~ZenSoo, ``{DeepSeek Has Rattled the AI Industry. Here’s a Look at Other Chinese AI Models},'' \emph{TIME}, Jan. 28 2025, accessed: Jan. 28, 2025. [Online]. Available: \url{https://time.com/7210521/deepseek-chinese-ai-models/}
\BIBentrySTDinterwordspacing

\bibitem{Yan2024}
\BIBentryALTinterwordspacing
{Yan et al.}, ``{Efficient and Accurate Prompt Optimization: The Benefit of Memory in Exemplar-Guided Reflection},'' \emph{arXiv}, vol. abs/2411.07446, Nov. 2024. [Online]. Available: \url{https://arxiv.org/pdf/2411.07446}
\BIBentrySTDinterwordspacing

\bibitem{Nie2024}
\BIBentryALTinterwordspacing
{Nie et al.}, ``{LSH-MoE: Communication-Efficient MoE Training via Locality-Sensitive Hashing},'' \emph{arXiv}, vol. abs/2411.08446, Nov. 2024. [Online]. Available: \url{https://arxiv.org/abs/2411.08446}
\BIBentrySTDinterwordspacing

\bibitem{AIbase2025}
{AIbase}, ``{iFlytek Releases the Xunfei Spark Deep Reasoning Model X1},'' Available at: \url{https://www.aibase.com/news/14723}, 2025, accessed: Jan. 28, 2025.

\bibitem{Kimi2025}
\BIBentryALTinterwordspacing
{Kimi Team}, ``{Kimi k1.5},'' GitHub repository, 2025, accessed: Jan. 28, 2025. [Online]. Available: \url{https://github.com/MoonshotAI/Kimi-k1.5}
\BIBentrySTDinterwordspacing

\bibitem{KimiTeam2025}
\BIBentryALTinterwordspacing
{KimiTeam et al.}, ``{Kimi k1.5: Scaling Reinforcement Learning with LLMs},'' \emph{arXiv}, vol. abs/2501.12599, Jan. 22 2025. [Online]. Available: \url{https://arxiv.org/abs/2501.12599}
\BIBentrySTDinterwordspacing

\bibitem{Ashley2025}
\BIBentryALTinterwordspacing
Ashley, ``{Kimi k1.5: How China’s New AI Powerhouse is Redefining Multimodal Reasoning and Beating OpenAI’s o1},'' \emph{Medium}, 2025, accessed: Jan. 28, 2025. [Online]. Available: \url{https://medium.com/@ashinno43/kimi-k1-5-how-this-next-gen-ai-model-is-revolutionizing-multimodal-reasoning-with-reinforcement-e06fbd64c12c}
\BIBentrySTDinterwordspacing

\bibitem{Qwen2025}
\BIBentryALTinterwordspacing
Qwen, ``{Qwen2.5-VL}.'' [Online]. Available: \url{https://github.com/QwenLM/Qwen2.5-VL/blob/main/README.md}
\BIBentrySTDinterwordspacing

\bibitem{OpenAIDeepResearch2025}
\BIBentryALTinterwordspacing
OpenAI, ``{Introducing Deep Research}.'' [Online]. Available: \url{https://openai.com/index/introducing-deep-research/}
\BIBentrySTDinterwordspacing

\bibitem{Sweney2025}
\BIBentryALTinterwordspacing
M.~Sweney and D.~Milmo, ``{OpenAI 'reviewing' allegations that its AI models were used to make DeepSeek},'' \emph{The Guardian}, Jan. 29 2025, accessed: Feb. 3, 2025. [Online]. Available: \url{https://www.theguardian.com/technology/2025/jan/29/openai-chatgpt-deepseek-china-us-ai-models}
\BIBentrySTDinterwordspacing

\bibitem{OpenAI2025}
{OpenAI}, ``{OpenAI o3-mini},'' Available at: \url{https://openai.com/index/openai-o3-mini/}, Jan. 31 2025.

\bibitem{Mulligan2025}
\BIBentryALTinterwordspacing
S.~J. Mulligan, ``{OpenAI Releases Its New o3-mini Reasoning Model for Free},'' \emph{MIT Technology Review}, Jan. 31 2025, accessed: Feb. 3, 2025. [Online]. Available: \url{https://www.technologyreview.com/2025/01/31/1110757/openai-makes-its-reasoning-model-for-free/}
\BIBentrySTDinterwordspacing

\bibitem{Jamali2025}
\BIBentryALTinterwordspacing
L.~Jamali, ``{China's DeepSeek AI Shakes Industry and Dents America's Swagger},'' \emph{BBC News}, Jan. 28 2025, accessed: Feb. 3, 2025. [Online]. Available: \url{https://www.bbc.co.uk/news/articles/cd643wx888qo}
\BIBentrySTDinterwordspacing

\bibitem{Wikipedia2025}
{Wikipedia}, ``{CHIPS and Science Act},'' Available at: \url{https://en.wikipedia.org/wiki/CHIPS_and_Science_Act}, 2025, accessed: Jan. 28, 2025.

\bibitem{Ng2025}
\BIBentryALTinterwordspacing
N.~Ng, B.~Drenon, T.~Gerken, and M.~Cieslak, ``{DeepSeek: The Chinese AI App That Has the World Talking},'' \emph{BBC News}, Jan. 27 2025, accessed: Jan. 27, 2025. [Online]. Available: \url{https://www.bbc.co.uk/news/articles/c5yv5976z9po}
\BIBentrySTDinterwordspacing

\bibitem{DeepSeekAI2025_HF}
\BIBentryALTinterwordspacing
{DeepSeek-AI}, ``{DeepSeek},'' Hugging Face, 2025, accessed: Jan. 27, 2025. [Online]. Available: \url{https://huggingface.co/deepseek-ai}
\BIBentrySTDinterwordspacing

\bibitem{Ollama2025}
{Ollama}, ``{deepseek-r1},'' Available at: \url{https://ollama.com/library/deepseek-r1}, 2025, accessed: Jan. 27, 2025.

\bibitem{Kellog2025}
\BIBentryALTinterwordspacing
T.~Kellog, ``{Someone on X Claims to Have Jailbroken R1 by Invoking the Name of Pliny, a Renowned LLM Jailbreaker},'' BlueSky, Jan. 24 2025, accessed: Jan. 27, 2025. [Online]. Available: \url{https://bsky.app/profile/timkellogg.me/post/3lgj25q42w22h}
\BIBentrySTDinterwordspacing

\bibitem{Martin2025}
\BIBentryALTinterwordspacing
{Martin et al.}, ``{DeepSh*t: Exposing the Security Risks of DeepSeek-r1},'' \emph{Hidden Layer}, Jan. 30 2025, accessed: Feb. 1, 2025. [Online]. Available: \url{https://hiddenlayer.com/innovation-hub/deepsht-exposing-the-security-risks-of-deepseek-r1/}
\BIBentrySTDinterwordspacing

\bibitem{Wilhoit2025}
\BIBentryALTinterwordspacing
K.~Wilhoit, ``{Recent Jailbreaks Demonstrate Emerging Threat to DeepSeek},'' \emph{Palo Alto Networks}, Jan. 30 2025, accessed: Feb. 1, 2025. [Online]. Available: \url{https://unit42.paloaltonetworks.com/jailbreaking-deepseek-three-techniques/}
\BIBentrySTDinterwordspacing

\bibitem{Thompson2025}
\BIBentryALTinterwordspacing
B.~Thompson, ``{DeepSeek FAQ},'' \emph{Stratechery}, Jan. 27 2025, accessed: Jan. 28, 2025. [Online]. Available: \url{https://stratechery.com/2025/deepseek-faq/}
\BIBentrySTDinterwordspacing

\bibitem{Kimmonismus2025}
\BIBentryALTinterwordspacing
@kimmonismus, ``{Billionaire and Scale AI CEO Alexandr Wang: DeepSeek Has About 50,000 NVIDIA H100s That They Can’t Talk About Because of the US Export Controls That Are in Place},'' X (formerly Twitter), Jan. 24 2025, accessed: Feb. 3, 2025. [Online]. Available: \url{https://x.com/kimmonismus/status/1882824571281436713}
\BIBentrySTDinterwordspacing

\bibitem{Dibya2025}
\BIBentryALTinterwordspacing
@its\_dibya, ``{With R1, a Lot of People Have Been Asking How Come We Didn't Discover This 2 Years Ago?}'' X (formerly Twitter), Jan. 26 2025, accessed: Feb. 3, 2025. [Online]. Available: \url{https://x.com/its_dibya/status/1883595705736163727}
\BIBentrySTDinterwordspacing

\bibitem{JiayiPirate2025}
\BIBentryALTinterwordspacing
@jiayi\_pirate, ``{The Specific RL Alg Doesn’t Matter Much…},'' X (formerly Twitter), Jan. 24 2025, accessed: Feb. 3, 2025. [Online]. Available: \url{https://x.com/jiayi_pirate/status/1882839504899420517}
\BIBentrySTDinterwordspacing

\bibitem{Janakiram2025}
\BIBentryALTinterwordspacing
J.~MSV, ``{All About DeepSeek – The Chinese AI Startup Challenging US Big Tech},'' \emph{Forbes}, Jan. 26 2025, accessed: Feb. 3, 2025. [Online]. Available: \url{https://www.forbes.com/sites/janakirammsv/2025/01/26/all-about-deepseekthe-chinese-ai-startup-challenging-the-us-big-tech}
\BIBentrySTDinterwordspacing

\bibitem{Stargate2025}
\BIBentryALTinterwordspacing
OpenAI, ``{Announcing The Stargate Project},'' \emph{{OpenAI Blog}}, Jan. 21 2025, accessed: Feb. 3 2025. [Online]. Available: \url{https://openai.com/index/announcing-the-stargate-project/}
\BIBentrySTDinterwordspacing

\bibitem{deSilvaFraser2025}
\BIBentryALTinterwordspacing
J.~de~Silva and G.~Fraser, ``{OpenAI Says Chinese Rivals Using Its Work for Their AI Apps},'' \emph{BBC News}, 2025, accessed: Feb. 3, 2025. [Online]. Available: \url{https://www.bbc.co.uk/news/articles/c9vm1m8wpr9o}
\BIBentrySTDinterwordspacing

\bibitem{Gerken2025}
\BIBentryALTinterwordspacing
T.~Gerken, ``{Be Careful with DeepSeek Australia Says – So Is It Safe to Use?}'' \emph{BBC News}, 2025, accessed: Jan. 28, 2025. [Online]. Available: \url{https://www.bbc.co.uk/news/articles/cx2k7r5nrvpo}
\BIBentrySTDinterwordspacing

\bibitem{Doffman2025}
\BIBentryALTinterwordspacing
Z.~Doffman, ``{New DeepSeek Warning — Do You Need To Delete Your iPhone, Android App?}'' \emph{Forbes}, Jan. 30 2025, accessed: Feb. 3, 2025. [Online]. Available: \url{https://www.forbes.com/sites/zakdoffman/2025/01/30/new-deepseek-warning-do-you-need-to-delete-your-iphone-android-app/}
\BIBentrySTDinterwordspacing

\bibitem{Pollina2025}
\BIBentryALTinterwordspacing
E.~Pollina, ``{DeepSeek blocked on Apple and Google app stores in Italy},'' \emph{Reuters}, Jan. 29 2025, accessed: Feb 3, 2025. [Online]. Available: \url{https://www.reuters.com/technology/deepseek-app-unavailable-apple-google-app-stores-italy-2025-01-29/}
\BIBentrySTDinterwordspacing

\bibitem{WizNagli2025}
\BIBentryALTinterwordspacing
G.~Nagli, ``{Wiz Research Uncovers Exposed DeepSeek Database Leaking Sensitive Information, Including Chat History},'' Jan. 29 2025, accessed: Feb 3, 2025. [Online]. Available: \url{https://www.wiz.io/blog/wiz-research-uncovers-exposed-deepseek-database-leak}
\BIBentrySTDinterwordspacing

\bibitem{Macaulay2025}
\BIBentryALTinterwordspacing
T.~Macaulay, ``{European AI alliance unveils LLM alternative to Silicon Valley and DeepSeek},'' \emph{The Next Web}, Feb. 3 2025, accessed: Feb. 3, 2025. [Online]. Available: \url{https://thenextweb.com/news/european-ai-alliance-openeurollm-challenges-us-china}
\BIBentrySTDinterwordspacing

\end{thebibliography}

\end{document}